\documentclass{article} 

\usepackage{iclr2025_conference,times}

\usepackage{amsmath,amsfonts,bm}

\def\eqref#1{equation~\ref{#1}}

\def\1{\bm{1}}

\DeclareMathAlphabet{\mathsfit}{\encodingdefault}{\sfdefault}{m}{sl}
\SetMathAlphabet{\mathsfit}{bold}{\encodingdefault}{\sfdefault}{bx}{n}

\usepackage[T1]{fontenc}
\usepackage[utf8]{inputenc}
\usepackage{float}
\usepackage{hyperref}
\usepackage{url}
\usepackage[table]{xcolor}
\usepackage{tabularx}
\usepackage{hyperref}
\usepackage{graphicx}
\usepackage{xcolor}
\usepackage{soul}

\sethlcolor{red} 

\usepackage{hyperref}
\usepackage{xr-hyper}

\makeatletter
\newcommand*{\addFileDependency}[1]{%
  \typeout{(#1)}%
  \@addtofilelist{#1}%
  \IfFileExists{#1}{}{\typeout{No file #1.}}%
}
\makeatother

\newcommand*{\myexternaldocument}[1]{%
    \externaldocument{#1}%
    \addFileDependency{#1.tex}%
    \addFileDependency{#1.aux}%
}

\myexternaldocument{supplementary_information}

\title{Conditional Diffusion Models for \\ Energy-Efficient Driving}

\author{Hemanth Neelgund Ramesh \\
Department of Mechanical Engineering, University of Washington \\
\texttt{hemnr31@uw.edu}
\AND
Andre Snoeck\thanks{Corresponding authors.\\
This work was conducted as part of the UW+Amazon Science Hub collaboration between the University of Washington and Amazon.}
\ \& Chyi-Fu Hong \\
Amazon Science \\
\texttt{\{snoeckas,chyifu\}@amazon.com}
\AND
Shijing Sun$^*$ \\
Department of Mechanical Engineering, University of Washington \\
Department of Materials Science and Metallurgy, University of Cambridge \\
\texttt{shijing@uw.edu}
}
\iclrfinalcopy
\begin{document}
\maketitle

\begin{abstract}
Electrification of commercial delivery fleets is shifting fleet routing from distance- and time-based optimization toward energy-aware decision-making. Existing sequence models primarily provide deterministic point estimates or limited uncertainty summaries, which do not capture the range of plausible energy-consumption trajectories required for operational decision-making. In this work, we introduce a conditional diffusion framework that generates EV battery-current profiles conditioned on route features such as vehicle velocity and ambient temperature. The model combines a latent conditioning encoder with a temporal 1D U-Net denoising backbone that enables trip-related conditions to be mapped into a shared representation and guides the reverse diffusion process. We evaluate the framework on an open-access commercial EV telemetry dataset containing ~ 12k trips from 9 vehicles. The proposed latent-conditioned diffusion model generates realistic current trajectories that capture both the dominant temporal envelope and sharp transient events. The model achieves a Wasserstein distance of 0.0029 between generated and measured current distributions below the real vs real reference distance of 0.0085 indicating that generated samples lie within the empirical variability of the test set. We further demonstrate that learned latent conditioning substantially improves performance over direct condition injection, reducing the Wasserstein distance by 89.1\% and MAE by 52.8\%. This work demonstrates a generative modeling framework for characterizing EV energy consumption under real-world operating conditions, providing an essential foundation for uncertainty-aware fleet planning in large-scale operational settings.
\end{abstract}


\section{Introduction}

The electrification of last- and middle-mile delivery fleets is accelerating across major logistics networks, with operators such as Amazon, UPS, and FedEx deploying electric vehicles at operational scale~\cite{Amazon2025SustainabilityReport, amazon_ev, fedex_ev, ups_ev}. This transition fundamentally changes the routing problem. Traditional optimization methods developed for gasoline vehicles often treat distance or travel time as reliable proxies for operating cost. However, for electric vehicles, these metrics are insufficient; energy consumption uncertainty has become a primary operational constraint~\cite{Andre2024}. Consequently, minimizing energy consumption and ensuring that trips can be completed within the available charge are now the primary objectives for fleet operators.

Accurately modeling this energy demand is challenging because electric vehicle energy consumption is highly context-dependent~\cite{AlWreikat2021, Zhang2020ap, Liu2018ap, Braun2018, Fiori2019}. Variations in road grade, ambient temperature, payload mass, and traffic dynamics can substantially alter energy requirements across otherwise similar trips. Therefore, effective modeling must integrate these heterogeneous data modalities, including dynamic features like velocity and its variations across terrains, alongside weather conditions. Furthermore, specific vehicle identity and battery chemistry dictate distinct discharge characteristics. The inherent stochasticity of driving behavior across different operational scenarios adds an additional layer of complexity that must be captured to reflect real-world operating conditions.

A growing body of literature on electric vehicle energy prediction has increasingly leveraged data-driven sequence models to estimate route-level energy demand from driving dynamics, vehicle characteristics, and environmental conditions~\cite{Hannan2021, Zhang2024review, Zhu2025ieee, Hussain2025, Yang2020, DeCauwer2017}. For instance, Feng et al.~\cite{Feng_energy_2024} proposed a hybrid LSTM-Transformer framework that combines recurrent temporal encoding with attention-based sequence modeling to predict electric vehicle energy consumption profiles. By incorporating multidimensional inputs including vehicle type, driving style, road conditions, and ambient factors, their model showed improved state-of-charge (SOC) prediction accuracy over standalone LSTM and multivariate regression baselines. Similarly, in the context of commercial fleet operations, Snoeck et al.~\cite{Andre2024} evaluated feed-forward networks, recurrent neural networks, and a Route Energy Transformer for last-mile delivery energy estimation. Their findings established that sequence-based architectures consistently outperform traditional distance-based heuristics and static physics models by natively capturing route-segment structure and delivery-specific operational dynamics. However, when trained with standard regression losses, these deterministic sequence models fundamentally predict the mean or median energy demand, thereby suppressing the operational variability critical for route-feasibility assessments. 

To address this limitation, prior work has introduced uncertainty-aware extensions, including quantile regression, prediction intervals, Bayesian neural networks, and ensemble-based methods~\cite{Hussain2025, khiari2023, Mahajan2024, Ozkan2024, Zhu2025}. For example, Petkevičius et al.~\cite{Petkevicius2021} developed probabilistic deep learning methods for electric vehicle energy-use prediction, where several models predicted distributional parameters rather than a single point estimate. Maity et al.~\cite{Maity_data-driven_2023} also proposed a probabilistic neural-network framework that incorporates model uncertainty using Monte Carlo approximation for trip-level energy demand. While these approaches emphasize that EV routing and range estimation benefit from uncertainty-aware predictions, sampling diverse, physically plausible energy-consumption trajectories under the same operational context requires a more flexible framework. This gap motivates the use of conditional generative models to directly characterize the full spread of possible energy outcomes.

Time-series generative models offer a principled approach to model energy consumption or current profiles by learning to sample trajectories conditioned on high-dimensional trip contexts. Among these, diffusion models are uniquely suited for this setting because they learn conditional distributions rather than collapsing into a single averaged trajectory. Recent literature has demonstrated the efficacy of these models across probabilistic time-series forecasting, imputation, and generation tasks~\cite{Yang2026, Su2025, meijer2024}. For example, Rasul et al.~\cite{Rasul_autoregressive_2021} introduced TimeGrad, an denoising diffusion model designed for multivariate probabilistic forecasting. Score-based approaches have also been studied, such as Conditional Score-based Diffusion Models for Imputation (CSDI), proposed by Tashiro et al.~\cite{Tashiro2021}, perform probabilistic time-series imputation by treating observed values as conditioning information through masking. More recently, the importance of conditioning on heterogeneous inputs has been demonstrated explicitly. The Time Weaver model by Narasimhan et al.~\cite{time-weaver} introduced a conditional diffusion framework that jointly encodes categorical, continuous, and time-varying metadata and injects this representation into the reverse denoising process. Their results show that structured conditioning on mixed-modality inputs can substantially improve generation quality.

\begin{figure}
    \centering
    \includegraphics[width=1\linewidth]{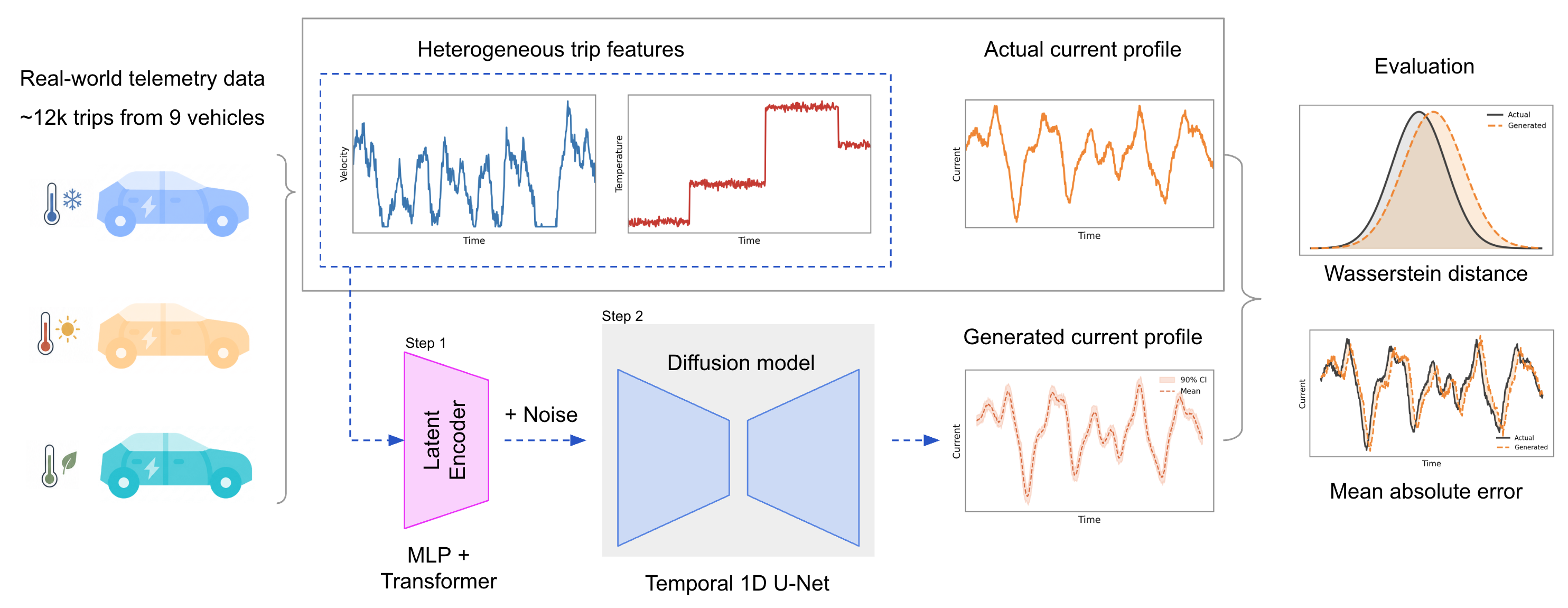}
    \caption{\textbf{Overview of the conditional diffusion framework for EV battery-current profile generation}. Route features (velocity and ambient temperature) from about 12,000 commercial trips are mapped by a latent encoder (MLP + Transformer) into a shared representation. This representation conditions a temporal 1D U-Net during reverse denoising to generate realistic battery-current time-series profiles evaluated against real-world trajectories.}
    \label{fig:framework}
\end{figure}

These architectural properties make diffusion models attractive for EV fleet energy modeling because they preserve temporal dependencies in sequential data while generating plausible trajectories under given input conditions. Here, we propose a conditional diffusion framework for generating EV battery-current profiles (figure~\ref{fig:framework} to support downstream applications such as state-of-charge (SoC) estimation, routing, and planning. Our approach generates battery-current time-series profiles conditioned on heterogeneous route features, including velocity and ambient temperature. To achieve this, we introduce a latent encoder that projects these input features into a shared representation, which is injected throughout the reverse denoising process of a 1D U-Net backbone. We evaluate the framework on a real-world commercial EV telemetry dataset comprising approximately 12,000 trips from nine vehicles and compare its distributional fidelity against deterministic and diffusion-based baselines. Overall, accurate generative modeling of battery-current profiles provides a foundation for uncertainty-aware planning in large-scale EV fleet operations.

\section{Methods}\label{methods}

\subsection*{Problem Formulation}

We formulate EV current modeling as a conditional time-series generation problem. For a trip of length $L$, the target is the battery current trajectory $\mathbf{i}_0=[i^{(1)},\ldots,i^{(L)}]^\top$. The conditioning input $\mathbf{c}$ represents trip-level operating context and may include dynamic signals, such as velocity sequences, as well as static or slowly varying descriptors, such as vehicle, route, terrain, payload, or weather-related temperature information when available. The learning objective is to model the conditional distribution $p_\theta(\mathbf{i}_0\mid \mathbf{c})$, rather than only a deterministic point estimate.

We approximate this distribution with a conditional diffusion model. Starting from Gaussian noise $\mathbf{i}_T\sim\mathcal{N}(\mathbf{0},\mathbf{I})$, the model iteratively denoises the signal over $T$ reverse-diffusion steps under the trip-specific conditioning $\mathbf{c}$ to generate a realistic current trajectory $\hat{\mathbf{i}}_0\sim p_\theta(\mathbf{i}_0\mid\mathbf{c})$. This distributional formulation allows multiple plausible current profiles to be sampled for the same trip context, which is useful for downstream SOC and energy-feasibility analyses where risk depends on the range of possible current and energy outcomes rather than a single predicted trajectory.

\subsection*{Data Sources and Preprocessing}

This study utilizes the open-access commercial-fleet electric vehicle (EV) dataset compiled by Rücker et al.~\cite{Rucker2024}, which comprises 1 Hz resolution battery and driving telemetry collected from nine commercial EVs. Each individual trip is interpolated to a fixed sequence length of $L = 512$ to preserve the relative temporal structure of each trip and ensure uniform input dimensions. Here, the battery current serves as the generation target, conditioned upon vehicle velocity and ambient temperature. After filtering out trips with excessive missing values, the primary dataset yielded a total of 11,945 valid trips. This sample was subsequently divided into training, validation, and testing partitions using a fixed 70/15/15 percentage split.  The data preprocessing protocols for both the datasets are detailed in Supplementary section S1 and Supplementary table S1.

\subsection*{Model Architecture}

\begin{figure}[h]
    \centering
    \includegraphics[width=1\linewidth]{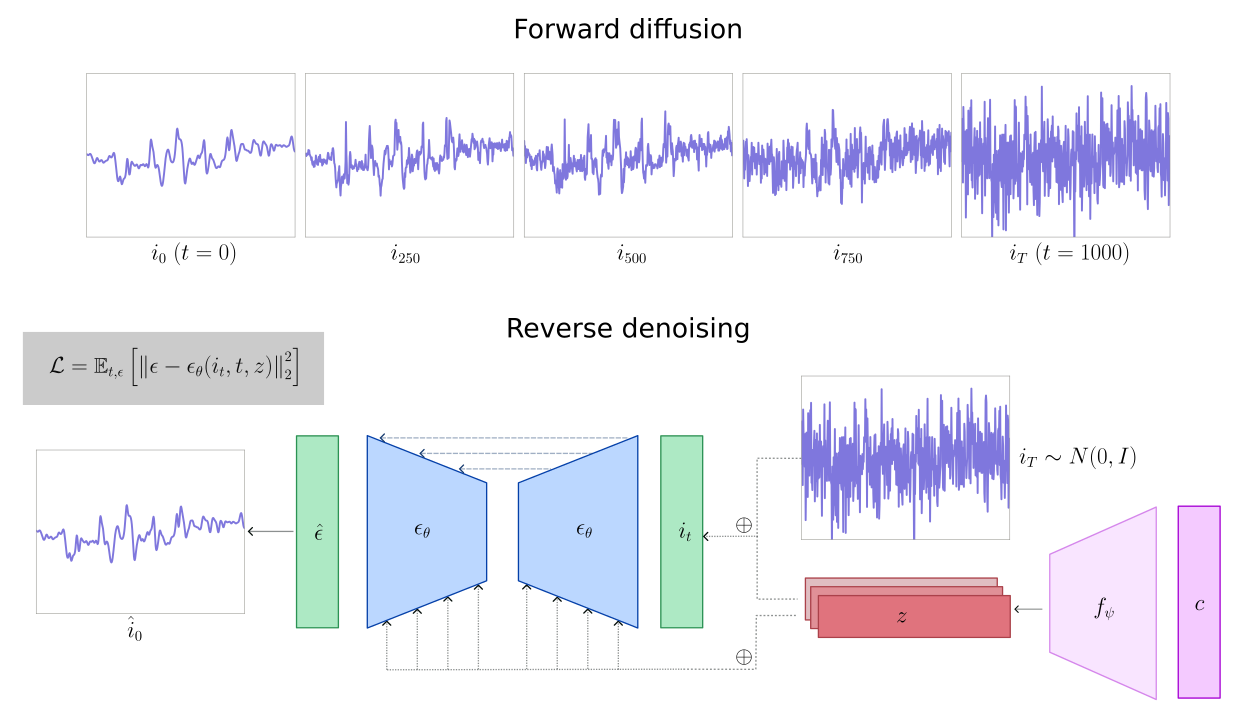}
    \caption{\textbf{Model architecture of the latent-conditioned diffusion framework.} The clean current trajectory \(i_0\) is progressively corrupted by Gaussian noise during the forward diffusion process to obtain \(i_t\). During reverse denoising, a temporal 1D U-Net predicts the added noise and reconstructs the current profile while being conditioned on the latent representation \(z=f_\psi(c)\), encoded from velocity and ambient temperature using the MLP-Transformer conditioning encoder.}
    \label{fig:model_arch}
\end{figure}

We model battery current trajectories using a two-stage conditional generation framework comprising of latent-conditioning encoder $f_{\psi}$ and a temporal 1D U-Net denoising diffusion model $\boldsymbol{\epsilon}_{\theta}$ (Fig.~\ref{fig:model_arch}). Here, the encoder $f_{\psi}$ and U-Net $\epsilon_{\theta}$ are trained jointly on single noise-prediction objective so that the conditioning representation is optimized directly for the generation task.  A full layer-by-layer specification in Supplementary table S2.

Recent conditional diffusion models for time-series tasks integrate heterogeneous conditioning signals, including categorical and continuous sequences, by using an explicit encoder to learn a shared representation before conditioning the denoising network. This separation is important in our setting because the conditioning variables have different temporal characteristics: vehicle velocity is a fast-varying sequence that captures second-scale driving dynamics, whereas ambient temperature provides a slowly varying environmental context. Direct concatenation would force the denoising backbone to simultaneously learn the temporal representation of the conditioning signals and the reverse diffusion dynamics. We therefore separate condition encoding from reverse denoising. 

The conditioning encoder $f_{\psi}$ maps the raw driving context $\mathbf{c} = [\mathbf{v}, \mathbf{T}] \in \mathbb{R}^{512 \times 2}$ to a contextualized latent representation through a channel-wise projection MLP, a sequence-wide Transformer encoder, and a linear compression head:

\begin{equation}
    \mathbf{z} = f_{\psi}(\mathbf{c}) \in \mathbb{R}^{512 \times 16}.
    \label{eq:latent_encoder}
\end{equation}

The latent condition $\mathbf{z}$ is concatenated channel-wise with the noisy current input at each reverse diffusion step. The configuration enables fixed-dimensional and preserves modality, so additional route and vehicle descriptors (terrain, payload, vehicle identity, traffic state) can be added through the encoder without modifying the denoising backbone.

The diffusion model is built on standard vanilla denoising diffusion probabilistic model (DDPM) framework~\cite{Ho2020}, in which a clean current trajectory is progressively corrupted by a fixed Gaussian forward process and a learned reverse process reconstructs the signal from noise. The forward transition and its closed-form marginal are provided in Supplementary Equations S2 and S3. 

The reverse denoising process is parameterized by a 1D temporal U-Net. Since EV current trajectories contain both slow and high-frequency transients due to acceleration and regenerative braking, the U-Net architecture provides a natural multi-resolution for this task. The encoder path compresses the sequence to capture broader temporal context, the decoder path reconstructs the signal at the original resolution, and skip connections preserve fine-scale information across matched resolutions. Dilated residual convolutional blocks expand the temporal receptive field without requiring excessive network depth~\cite{oord2016, kong2020}. Whereas anti-aliased BlurPool downsampling reduces aliasing during temporal compression~\cite{richzhang2019, Zou2020}. The complete U-Net architecture is summarized in Supplementary Table S2.

The encoder and U-Net are trained jointly using the standard noise-prediction objective,
\begin{equation}
    \mathcal{L}
    =
    \mathbb{E}_{\mathbf{i}_0, \mathbf{c}, t, \boldsymbol{\epsilon}}
    \left[
    \left\|
    \boldsymbol{\epsilon}
    -
    \boldsymbol{\epsilon}_{\theta}
    \bigl(\mathbf{i}_t, t, f_{\psi}(\mathbf{c})\bigr)
    \right\|_2^2
    \right],
    \label{eq:loss}
\end{equation}
where $\mathbf{i}_0$ is the clean current trajectory, $\mathbf{i}_t$ is the noisy trajectory at diffusion step $t$, and $\boldsymbol{\epsilon}$ is the Gaussian noise sampled during the forward process.

During inference, trajectories are sampled by initializing from Gaussian noise and applying the learned reverse process conditioned on $\mathbf{z}$. We use $T=1000$ diffusion steps and a cosine noise schedule~\cite{Nichol2021}, which corrupts the signal more gradually than a linear schedule and preserves transient structure for longer in the forward process. Training and sampling pseudocode are provided in Supplementary Algorithms S1 and S2.

\subsection*{Experiments}

We systematically evaluate the core architectural design choices by conducting experiments assessing latent representation learning, conditioning variable selection, diffusion noise scheduling, and model capacity. These experiments are summarized in Supplementary Table S3. The proposed architecture, Model E4, uses the latent encoder $f_{\psi}(\mathbf{v}, \mathbf{T})$ and a cosine noise schedule.

The experimental configurations are structured to evaluate specific components of the proposed architecture. To isolate the utility of the learned latent space, Model D0 bypasses the latent encoder $f_{\psi}$ and directly injects raw condition sequences into the U-Net. Furthermore, the relative importance of individual conditioning variables is evaluated through feature ablation, where Models D1 and D2 exclude velocity $(\mathbf{v})$ and ambient temperature $(\mathbf{T})$ from the conditioning vector, respectively.

Apart from architectural modifications, additional variants evaluate hyperparameter sensitivity by replacing the cosine schedule with a standard linear noise schedule and by varying the channel depth of the 1D U-Net. Finally, to compare generative performance against standard sequential modeling approaches, a deterministic Long Short-Term Memory (LSTM) network. The LSTM baseline represents a conventional point-prediction framework, whereas the proposed diffusion model captures distributional variability.

\subsection*{Evaluation Metrics}

We evaluate the generated trajectories using pointwise reconstruction errors (MAE and RMSE), spectral fidelity via short-time Fourier transform (STFT) distance, and distributional alignment via the 1-Wasserstein distance (WD). Together, these metrics provide collective evidence to compare the models.


\section{Results}

\begin{figure}
    \centering
    \includegraphics[width=1\linewidth]{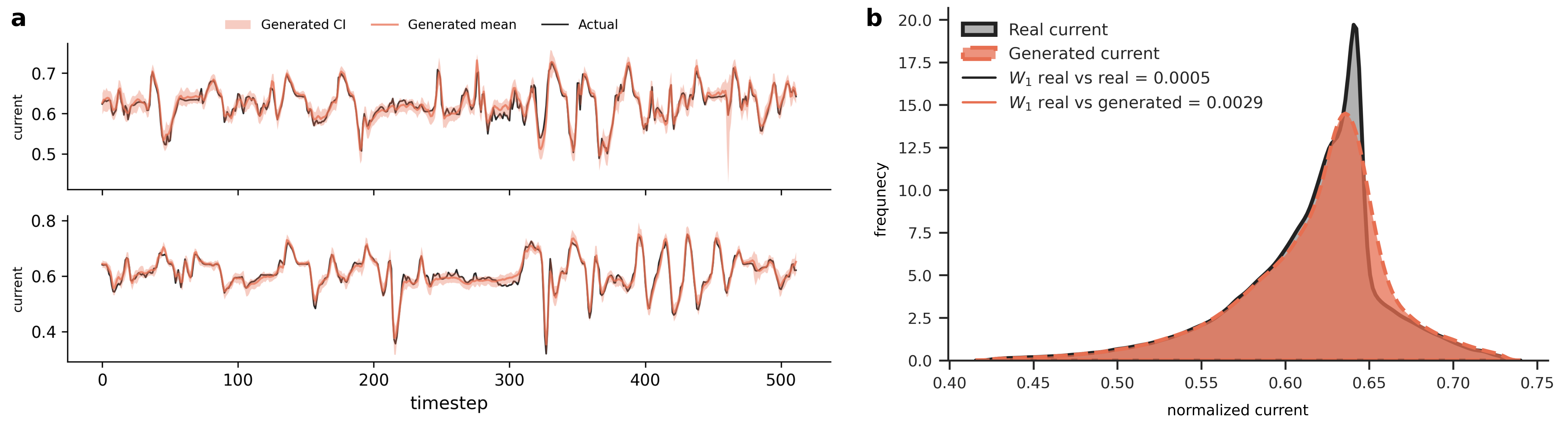}
    \caption{\textbf{Conditional diffusion model generates realistic battery current trajectories}. (a) Plot comparing generated trajectories (mean and 95\% confidence interval) vs actual current profiles for two held-out test trips for the proposed model. (b) Distribution of Wasserstein distance (WD) computed on a held-out test trips.}
    \label{fig:proposed_model_results}
\end{figure}

Figure~\ref{fig:proposed_model_results}a compares generated and actual measured battery-current trajectories for representative held-out trips from the primary test set using the proposed model (E4). The generated trajectories follow the dominant current envelope across the trip and capture sharp transients related to acceleration, deceleration, and regenerative-braking events. The shaded region shows the spread across 10 independent generated samples for the same conditioning input. This indicates that the model produces a distribution of plausible trajectories, including meaningful spread at peak amplitudes. proposed model captures the temporal structure of EV current demand under observed trip conditions while maintaining the uncertainty spread samples.

The distributional comparison in figure~\ref{fig:proposed_model_results}b provides a quantitative measure of generation quality. The proposed model gives a WD of 0.0029 between generated and actual (real) current values, compared with a real-versus-real reference value of 0.0085. Since the generated-versus-real distance is lower than the distance between two subsets of measured data, the generated current distribution lies within the empirical variability of the test set.

The denoising trajectory in Supplementary figure S2 provides a diagnostic view of the reverse sampling process. The heatmap shows large across-sample variance near the initial noise state, followed by a rapid reduction in variance as the model approaches the final generated current trajectory, \(x_0\). Most of the visible variance reduction has occurred by approximately \(t=250\), indicating that the model has already formed the dominant current structure by this stage. For \(t<250\), the variance is significantly low, suggesting that the remaining reverse steps mainly refine local fluctuations. This behavior is expected for conditional diffusion models which map Gaussian noise to structured data, particularly when using cosine noise schedule.

\begin{figure}[t]
    \centering
\includegraphics[width=1\linewidth]{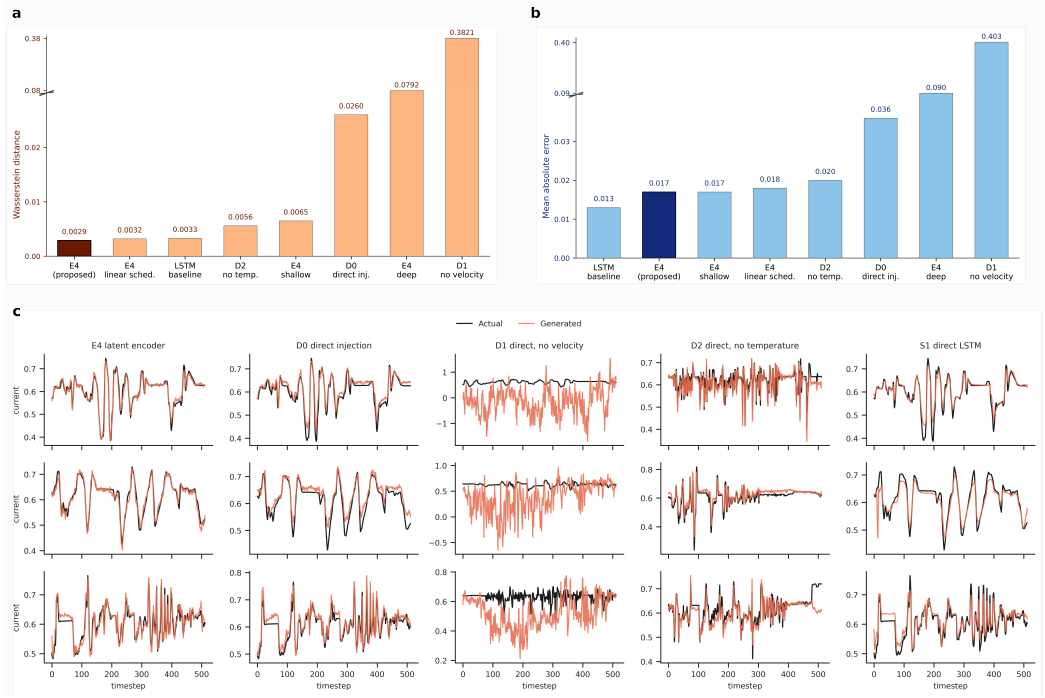}
    \caption{\textbf{Model comparison and ablation results.}
    (a,b)Wasserstein distance (WD) and Mean absolute error (MAE) for the selected model variants, shown in ascending order. The proposed E4 model is highlighted in dark. 
    (c) Generated and actual current trajectories for selected models on the same test sample. D1 and D2 receive only temperature or velocity conditioning, respectively.}
\label{fig:experiments_results}
\end{figure}

We compare the proposed model with the deterministic and diffusion-based variants described in section~\ref{methods}. Figure~\ref{fig:experiments_results}a,b reports the bar plot for WD and MAE for each model.The proposed E4 model achieves the best overall performance, ranking second in MAE behind only the LSTM. This is expected as LSTM is a deterministic sequence predictor optimized on a standard regression loss. In contrast, E4 is designed to generate plausible trajectories from a conditional distribution, resulting in a significantly better WD score.

Among the E4-based diffusion variants, the choice of noise schedule has a weaker influence on performance than the conditioning strategy or network capacity. The E4 linear-schedule variant performs close to the proposed E4 cosine-schedule model, suggesting that both schedules can effectively learn the denoising process for this dataset. However, E4 shows slightly better distributional fidelity due to smooth signal corruption caused by the cosine schedule. With respect to channel depth, the E4-shallow model gives MAE values comparable to the proposed E4 model. However, further increasing the U-Net channel depth degrades performance, as shown by the E4-deep model. Because the temporal U-Net already incorporates down sampling and residual blocks, increasing network depth simply over-smoothens the trajectories.

Figure~\ref{fig:experiments_results}a,b separates the effects of feature ablation and conditioning strategy. D0 uses the same conditioning variables as E4, but injects the raw condition sequence directly into the U-Net instead of passing it through the latent encoder. This results in approximately a two-fold increase in MAE and a nine-fold increase in WD relative to E4. Since the input conditions are same, the performance gap is due to the importance of the learned latent-conditioning pathway. Furthermore, removing velocity condition in D1 increases the error significantly, which indicates that velocity provides strong context needed to generate current trajectories. However, removing  temperature  condition in D2 shows a much smaller effect, which is reasonable given temperature is comparatively slow-varying than velocity.

Figure~\ref{fig:experiments_results}c compares actual and generated trajectories for selected models. The proposed E4 model and the LSTM baseline both follow the actual trajectory closely. D0 shows small deviations, further supporting the role of the latent encoder. The feature-ablation models, D1 and D2, show noticeable deviations from the actual trajectory, particularly when velocity is removed. The D2, which retains velocity as the conditioning input, still captures much of the overall trajectory structure. This trend is consistent with the earlier quantitative results in Fig.~\ref{fig:experiments_results}a,b.

\begin{figure}[t]
    \centering
    \includegraphics[width=1\linewidth]{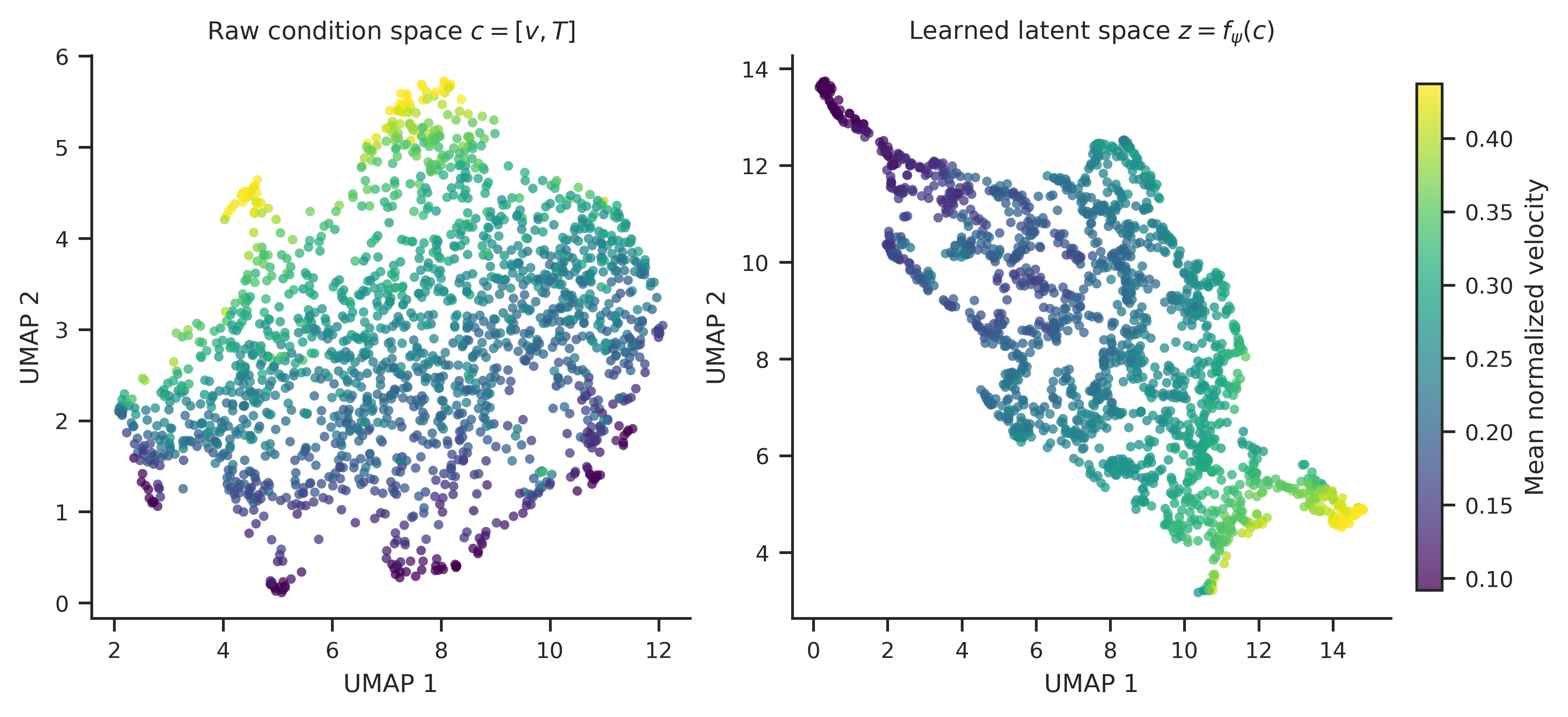}
    \caption{\textbf{Learned latent representation of operating conditions.}
    UMAP projections of the raw condition sequence $c=[v,T]$ (left) and the learned latent condition $z=f_\psi(c)$ (right) colored by mean normalized velocity.}
    \label{fig:umap}
\end{figure}

Figure~\ref{fig:umap} further underscores the importance of structurally encoding heterogeneous conditioning variables. We compare UMAP projections of the raw conditioning input, \(c=[v,T]\), and the learned latent condition, \(z=f_\psi(c)\), colored by mean normalized velocity. The raw conditioning space is comparatively diffuse and does not show a clear organization with respect to velocity. In contrast, the learned latent representation shows a more ordered structure, with samples arranged along a velocity-dependent trend.

This structured representation is important because velocity carries much of the information needed to generate current trajectories, as shown earlier. The visualization also explains the performance gap between the proposed E4 model and the direct-conditioning D0 model. Although both models receive same inputs, E4 first learns an intermediate latent representation before denoising, which yields an improved trajectory generation. These findings indicate that effective representation learning is essential for generating high-fidelity current trajectories.

\section{Discussion}

In this work, we model EV battery usage as a generative time-series task to capture the inherent stochasticity of real-world fleet operations. Deterministic sequence models, such as LSTMs, yield accurate point predictions but they collapse variability into a single averaged trajectory. On the other hand, our conditional diffusion model successfully learns the distribution, which enables sampling plausible current profiles for the same trip conditions. Ultimately, this approach supports downstream fleet applications, such as SOC estimation, battery health aware routing, and operational analysis, which depends on estimating accurate change consumption across fleets.

We developed a two-stage model architecture that couples a latent encoder with a U-Net diffusion backbone. The encoder projects raw, multi-modal driving context into a shared latent space, which is subsequently concatenated with the noisy current input at each reverse diffusion step. This decoupled design ensures the framework is highly extensible, allowing for the integration of additional conditioning channels (such as road grade, vehicle payload, fleet identity, or real-time traffic states) if available. By adding such route descriptors through a unified encoder pathway, diverse modalities can be fused into the latent representation without modifying the underlying denoising backbone. The encoder serves as a modular interface between complex environmental inputs and the core generative model.

Experimental results demonstrate that the proposed E4 architecture outperforms other diffusion-based (from the experiments) and significantly improves upon the LSTM in terms of WD. This This is because the model efficiently learns to generate a distribution of potential current trajectories for a single context, while also preserving sharp, highly transient current peaks. Among the input conditions, velocity appears as the dominant conditioning signal as it indirectly reflects driver behavior, traffic dynamics, and road topology, whereas ambient temperature provides only sparse, slowly varying context. We also show that samples organize in a highly structured manner with respect to velocity when passed through the encoder; consequently, the model achieves superior generation fidelity compared to direct condition injection. Furthermore, we observe that the choice of noise schedule has minimal influence on generation quality; instead, performance is heavily driven by the architecture itself, specifically the integration of residual connections and BlurPool layers. This also explains why increasing channel depth showed diminishing returns. 

Despite these promising results, there still exist some limitations that must be acknowledged. First, our framework relies on publicly available passenger-EV telemetry, which features a narrow set of conditions (primarily velocity and temperature). While richer, multi-sensor datasets exist, they are severely limited in scale, often containing only a few hundred trips~\cite{Steinstraeter2020, Zhang2022, Tesla2025}. However, the proposed latent encoder is structurally equipped to accommodate their multi-modal inputs into an organized latent representation. Second, the trajectories are generated at a fixed, interpolated length, meaning physical trip duration enters the model only implicitly through the conditioning channels. Finally, while the current framework demonstrates robust capabilities in generating realistic samples, distributional fidelity alone is insufficient for operational deployment. To fully demonstrate its value, the model must be evaluated in a closed-loop framework within an active routing and planning layer. 

Consequently, a natural next step is to extend the framework to accommodate variable-length trajectories natively. This can be achieved by integrating an adaptive projection layer prior to the diffusion process, or by transitioning from a U-Net to a Diffusion Transformer (DiT) backbone. Furthermore, while this work focuses on generating uni-variate battery current profiles, the framework can be expanded to multi-variate time-series generation (such as co-generating synchronized voltage and current profiles) to provide rich battery state information for energy-aware tasks. Ultimately, coupling these generative models with downstream optimization algorithms, such as reinforcement learning, will allow fleet operators to utilize the generated profiles across sequential segments, minimizing aggregate energy consumption and maximizing operational feasibility.


\section{Conclusion}
In this work, we developed a conditional diffusion model for generating time-series EV battery-current profiles. The proposed architecture combines a conditioning encoder, which maps velocity and ambient temperature into a structured latent representation, with a temporal 1D U-Net diffusion backbone that reconstructs realistic current trajectories through reverse denoising.

Our results show that the proposed E4 model achieves the strongest overall performance among the evaluated diffusion variants in terms of both Wasserstein distance and MAE. The LSTM model gives slightly lower MAE, this is expected because it is deterministic by design and optimized for pointwise prediction. The ablation studies further show that learned latent conditioning is important for improving generation quality, and that velocity is the dominant conditioning signal for battery-current profile generation.

We hope this approach to support future decision-making in energy-aware fleet operations by enabling route planners to evaluate energy outcomes under real-world driving conditions. Future work should emphasize the use of datasets with richer route descriptors, including detailed weather patterns, traffic and road dynamics, variable-length trips, multivariate battery signals, and integration with closed-loop routing or planning systems.


\subsubsection*{Data and Code Availability}
The commercial-fleet EV dataset used in this study is openly 
available from R\"{u}cker et al.~\cite{Rucker2024} at 
\url{https://publications.rwth-aachen.de/record/979878}. 
Code for data preprocessing and training the diffusion models 
is available at \url{https://github.com/nrhemanth/evdiff}.

\subsubsection*{Author Contributions}
H.N.R.: Conceptualization, Methodology, Software, Validation, 
Visualization, Data Curation, Writing - Original Draft, Writing 
- Review \& Editing. A.S. and C.-F.H.: Conceptualization, Methodology, Validation, Data Curation, Visualization, Writing - Review \& Editing. S.S.: Conceptualization, Methodology, Visualization, Validation, Data Curation, Writing - Review \& Editing, Funding Acquisition.

\subsubsection*{Acknowledgments}
This work was supported by the UW+Amazon Science Hub.

\subsection*{Declaration of Interests}
The authors declare no conflict of interest.

\subsection*{Declaration of AI-assisted Technologies}
During the preparation of this work, the authors used LLM tools (ChatGPT and Claude code) in writing process and to help resolve Python coding errors. After using this tool, the authors carefully reviewed and edited all content as needed, and take full responsibility for the final version of the publication.


\bibliographystyle{unsrtnat}
\bibliography{references}


\appendix
\newpage

\end{document}